\RequirePackage{fix-cm}
\documentclass[twocolumn]{svjour3}          % twocolumn
\smartqed  % flush right qed marks, e.g. at end of proof
\usepackage{graphicx}
\journalname{Applied Intelligence}
\usepackage{hyperref}

\usepackage[utf8]{inputenc}
\usepackage[misc]{ifsym} % Letter symbol; correct variant...?
\usepackage{siunitx}
\usepackage{subcaption}
\usepackage{pbox}
\newcommand*{\inst}[1][*]{\textsuperscript{#1}}
\begin{document}\sloppy
\title{Temperate Fish Detection and Classification: \\a Deep Learning based Approach}
\titlerunning{Temperate Fish Detection and Classification}
% If the paper title is too long for the running head, you can set
% an abbreviated paper title here
\authorrunning{K. Knausgård et al.}
\author{Kristian~Muri~Knausgård\inst[1]\textsuperscript{(\Letter)} \and Arne~Wiklund\inst[2] \and Tonje~Knutsen~Sørdalen\inst[3,4]\and Kim~Halvorsen\inst[3] \and Alf~Ring~Kleiven\inst[3] \and Lei~Jiao\inst[2] \and Morten~Goodwin\inst[2]}
\institute{\inst[1] Department of Engineering Sciences, University of Agder (UiA), 4879 Grimstad, Norway\and \inst[2] Centre for Artificial Intelligence Research, UiA, 4879 Grimstad, Norway \and \inst[3] Institute of Marine Research (IMR), His, Norway \and \inst[4] Department of Natural Sciences, Centre for Coastal Research (CCR) UiA, 4630 Kristiansand, Norway.
\email{kristianmk@ieee.org}}
\date{Received: date / Accepted: date}
% The correct dates will be entered by the editor
%
\maketitle % typeset the header of the contribution
\begin{abstract} % Springer: The abstract should briefly summarize the contents of the paper in 15--250 words.
A wide range of applications in marine ecology extensively uses underwater cameras. Still, to efficiently process the vast amount of data generated, we need to develop tools that can automatically detect and recognize species captured on film. Classifying fish species from videos and images in natural environments can be challenging because of noise and variation in illumination and the surrounding habitat. In this paper, we propose a two-step deep learning approach for the detection and classification of temperate fishes without pre-filtering. The first step is to detect each single fish in an image, independent of species and sex. For this purpose, we employ the You Only Look Once (YOLO) object detection technique. In the second step, we adopt a Convolutional Neural Network (CNN) with the Squeeze-and-Excitation (SE) architecture for classifying each fish in the image without pre-filtering. We apply transfer learning to overcome the limited training samples of temperate fishes and to improve the accuracy of the classification. This is done by training the object detection model with ImageNet and the fish classifier via a public dataset (Fish4Knowledge), whereupon both the object detection and classifier are updated with temperate fishes of interest. The weights obtained from pre-training are applied to post-training as a priori. Our solution achieves the state-of-the-art accuracy of 99.27\% on the pre-training. The percentage values for accuracy on the post-training are good; 83.68\% and 87.74\% with and without image augmentation, respectively, indicating that the solution is viable with a more extensive dataset.
% Removed to stay below 250 word limit:  In previous methods, a pre-filtering process is usually utilized, which relies on separating the fish from the background or sharpen the image by image processing techniques. However, this may remove useful information for classification and thus degrade the classification accuracy. 
%\todo[inline]{I understand that the plural of fish is fish. But multiple species of fish is fishes. Done by Lei}
%
%
\keywords{ Biometric Fish Classification \and Temperate Species\and Deep Learning \and Object Detection \and CNN \and Underwater Video}
\end{abstract}

\section{Introduction}\label{Introduction}
Coastal marine ecosystems provide habitats for spawning, nursing, and feeding for a diverse fish community. Due to the highly complex and dynamic nature of this environment, it is challenging to monitor and study ecological processes ~\cite{Perry2018Habitat,Weinstein2017AComputerVision}. High resolution underwater camera technologies have recently made it possible to obtain large volumes of observations from remote areas and allowed for better capture the species' cryptic behavior and changes in the environment \cite{PELLETIER201184}. Although comprehensive image and video data can be collected, the processing is of image data in ecological context is mostly manual and therefore very labor-intensive \cite{Lopez2020}. As a result, only a portion of the available recordings can be analyzed which is greatly limiting the potential advances that can be made from these data streams. Furthermore, the accuracy of human-based visual assessments are highly dependent on conditions of the underwater environment and taxonomic expertise in interpreting the data \cite{Francour1999ComparisonOfFishAbundanceEst}. Therefore, an objective analytical tool capable of processing image data fast and efficient is most welcomed by scientists and resource management.
\begin{figure*}[ht!]
    \centering
    \includegraphics[width=0.65\linewidth]{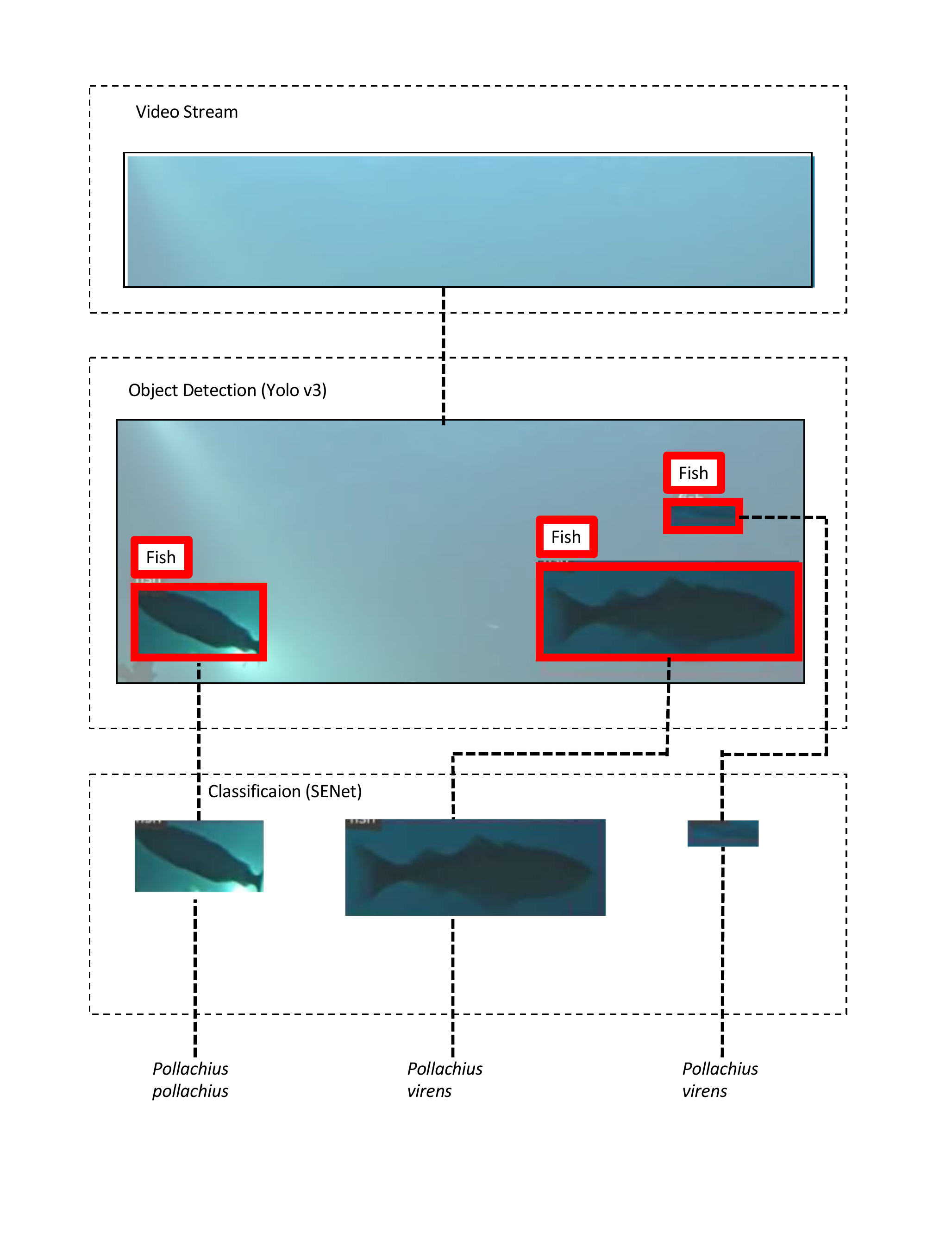}
    \caption{\label{Architecture}Architecture.}
\end{figure*}

To release the burden of manual processing, and to improve the classification accuracy, computer vision-based approaches have increasingly been employed in marine ecology analysis \cite{li2015fast,qin2016deepfish,jin2017deep}. For instance, a commercial product, CatchMeter \cite{white2006automated}, composed by a lightbox with a camera, offers classification of fish and length estimates. Here, fish are classified by evaluating a threshold based on a contour detection in the images with a very high classification accuracy of 98.8\%. However, the fish are photographed in a pre-determined and controlled environment, which hinders applying the approach in the wild. The CatchMeter version described in \cite{white2006automated} does not make use of any AI or machine learning techniques. In natural underwater environments, any classification task is challenged by diversity in background complexity, turbidity and light propagation as the water deepens.

A specific Convolutional Neural Network (CNN) called Fast R-CNN has been applied for object detection to extract the fish from images taken in natural environment and actively ignoring background noise \cite{li2015fast}. In this approach, an AlexNet \cite{krizhevsky2012imagenet} is pre-trained on the ImageNet~\cite{deng2009imagenet} database and modified to train on a subset of the Fish4Knowledge dataset \cite{fish4knowledge}. In the final step, the Fast R-CNN takes the pre-trained weights and the region proposals made by AlexNet as inputs, and achieves a mean average precision of 81.4\%. In another approach \cite{jin2017deep}, pre-training is applied to a CNN similar to AlexNet, which has three fully-connected layers and five convolutional layers. Pre-training is carried out using 1000 images from 1000 categories in the ImageNet dataset and the learned weights are utilized by a CNN after adapting it to the Fish4Knowledge dataset. Post-training is then performed with 50 images per category and 10 categories from the Fish4Knowledge. The images from Fish4Knowledge are pre-processed using image de-noising and accuracy achieved on 1420 test images is 85.08\%.

The highest reported accuracy for Fish4Knowledge in the literature so far is 98.64\%, which was achieved by firstly utilizing filters to the original images to extract the shape of the fish and remove the background, and then employing a CNN with a Support Vector Machine (SVM) for classification~\cite{qin2016deepfish}. That approach is named DeepFish, which has three standard convolution layers and three fully-connected layers. One common feature of previous solutions is that they usually adopt a pre-processing procedure for the images in order to remove the noise in the targeted image as much as possible, and particularly to outline the contour of the fish~\cite{jin2017deep,qin2016deepfish}. Although this method can improve the system performance, the procedure of the pre-process must be carefully tuned, as it may remove useful information and result in a negative performance impact. Understandably, different species may have distinct nature of living environment, reflected in the background. Intentionally removing the background of the species in the pre-processing may therefore eliminate useful information. To make use of information from the background as much as possible and at the same time to keep the results not influenced by background noise, we need to employ a robust approach that can tolerate noise and accommodate diversity in classification.

In previous work on fish detection, Liu et al. (2018) have presented an online fish tracking system using YOLO and parallel correlation filters, and included detection and categorization in an end-to-end approach~\cite{8604658}. Similar work is carried out by Xu et al. (2018) who trained a YOLO architecture aimed at detecting a variety of fish species with three very different datasets, obtaining a mean average precision score of 0.5392~\cite{underwaterfishwaterpower}. Pedersen et al. (2019) extended their work to include marine mammals as well as fish and used the same YOLO techniques~\cite{PedersenDetectionOfMarineAnimals}. Common for all of these approaches is that they trained their network end-to-end.

In this paper, we propose a different method, namely a separate deep learning-based approach for temperate fish detection and classification. In more detail, we have used images, and videos taken by underwater cameras in natural environments, employed YOLOv3~\cite{yolov3techreport} for fish detection, and explored CNN using the most recent SE architecture for classification. Because it is common to have multiple species in the same frame, the YOLO algorithm was used for fish detection, and once detected, the algorithm classified the fish to its particular species. Because the Fish4Knowledge dataset is limited to tropical fish species, for the training samples in the classification phase, we collected a new dataset of temperate fish species for this study. Our approach for classification was to train the network on the Fish4Knowledge dataset in order to learn generic features of fish, a step called pre-training. The learned weights were then used as a starting point for further training on the newly collected dataset containing images of temperate fish species, called post-training. This two-step training process is known as transfer learning~\cite{YosinskiTransferLearning}. Note that the proposed approach requires no pre-processing of images, except re-sizing to the appropriate input size for the network. To the best of our knowledge, the adopted techniques have not been applied to temperate fish detection and classification in previous work. 

The remainder of the paper is organized as follows: Section~\ref{method} describes the datasets adopted for the training process. Section~\ref{structure} presents a detailed network structure and configurations. In Section~\ref{experimentalResults}, the experimental results for the deep learning approach is illustrated and discussed, before the work is concluded in the last section. An abridged version of this article is published in~\cite{olsvik2019}.
\section{Datasets and Deep Learning Approaches}\label{method}
Fig.~\ref{Architecture} presents the overall architecture of our approach. First, a video stream is sent into an object detection component, which is a YOLOv3 CNN. YOLOv3 is pre-trained on ImageNet and fine-tuned for detecting temperate fish species using a custom dataset. This component detects the presence of fish in a single video frame, and moves the rectangular subframes with fish to a classification component built on a CNN-SENet structure. The latter categorizes the fish species, and the overall architecture is thus able to count the number of fish belonging to each species in each frame. The components are trained individually -- the fish detection training is completely independent of the fish species classification training. This separation has two main advantages. First, the training data for categorization and object detection is allowed to be separate. It is tedious to outline every single fish in a video stream. Since object detection of fish requires less data than classification of fish species, the biologists can spend their time mostly on specialist work like categorization, rather than outlining objects. Second, detecting the presence of fish is a more straightforward problem than categorizing species, which means that we can prioritize resources accordingly.

\begin{figure*}[ht]
\begin{subfigure}{.25\textwidth}
  \centering
  \includegraphics[width=.9\linewidth]{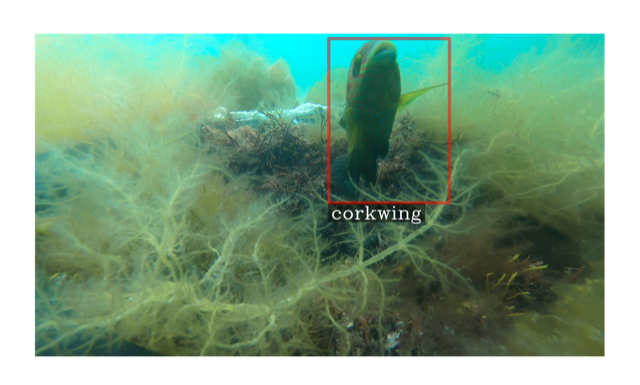}
\end{subfigure}%
\begin{subfigure}{.25\textwidth}
  \centering
  \includegraphics[width=.9\linewidth]{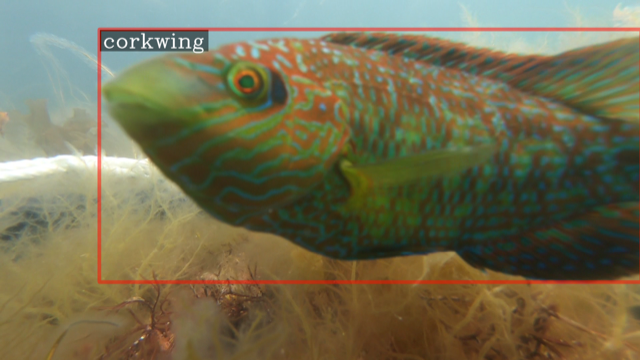}
\end{subfigure}%
\begin{subfigure}{.25\textwidth}
  \centering
  \includegraphics[width=.9\linewidth]{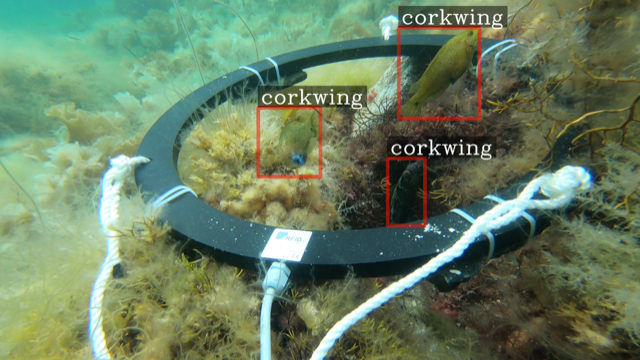}
\end{subfigure}%
\begin{subfigure}{.25\textwidth}
  \centering
  \includegraphics[width=.9\linewidth]{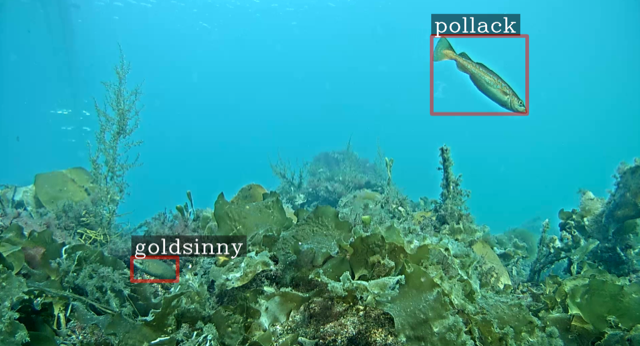}
\end{subfigure}%
\vspace{0.4cm}
\\
%\end{figure*}
%\begin{figure*}
\begin{subfigure}{.25\textwidth}
  \centering
  \includegraphics[width=.9\linewidth]{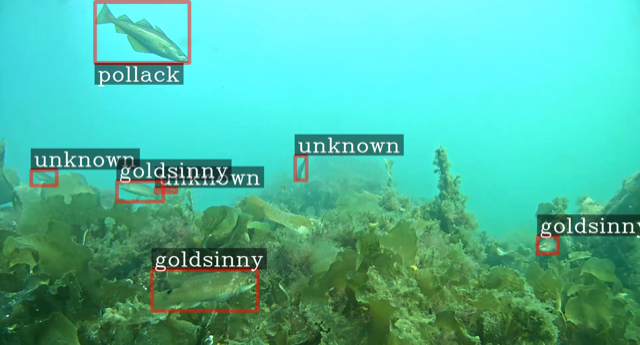}
\end{subfigure}%
\begin{subfigure}{.25\textwidth}
  \centering
  \includegraphics[width=.9\linewidth]{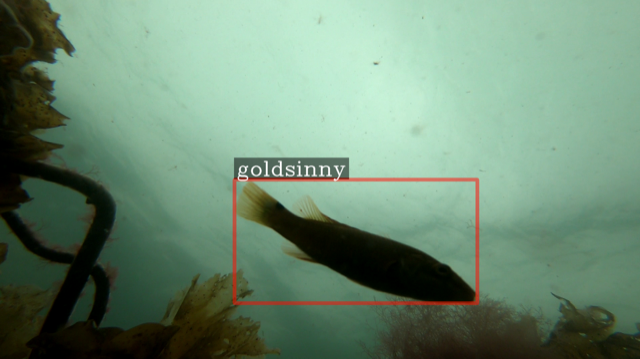}
\end{subfigure}%
\begin{subfigure}{.25\textwidth}
  \centering
  \includegraphics[width=.9\linewidth]{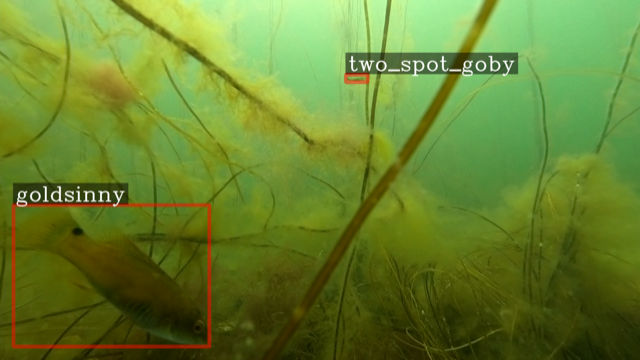}
\end{subfigure}%
\begin{subfigure}{.25\textwidth}
  \centering
  \includegraphics[width=.9\linewidth]{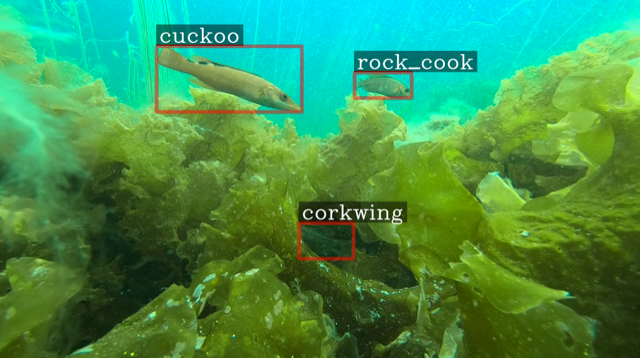}
\end{subfigure}%
\vspace{0.4cm}
\\
%\end{figure*}
%\begin{figure*}
\begin{subfigure}{.25\textwidth}
  \centering
  \includegraphics[width=.9\linewidth]{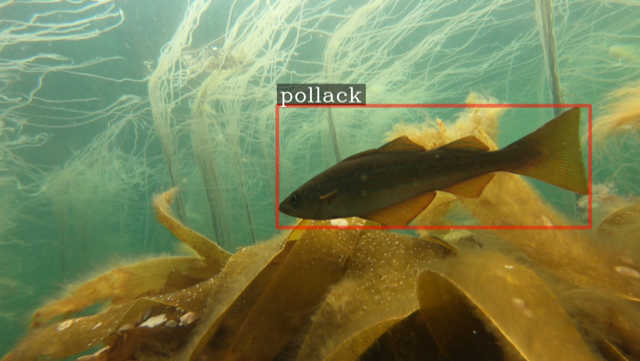}
\end{subfigure}%
\begin{subfigure}{.25\textwidth}
  \centering
  \includegraphics[width=.9\linewidth]{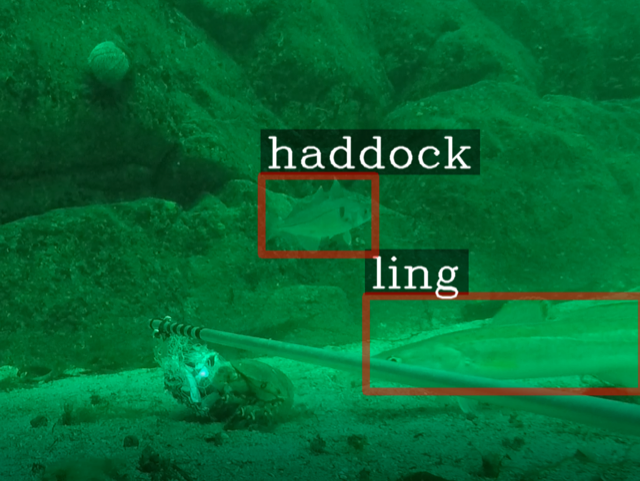}
\end{subfigure}%
\begin{subfigure}{.25\textwidth}
  \centering
  \includegraphics[width=.9\linewidth]{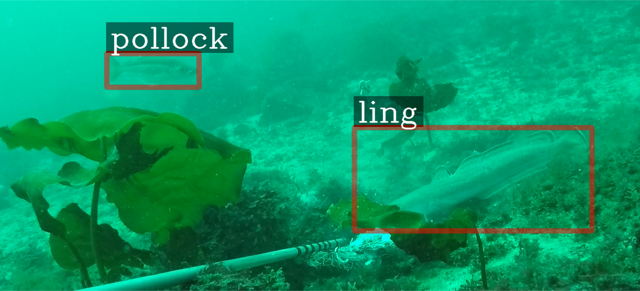}
\end{subfigure}%
\begin{subfigure}{.25\textwidth}
  \centering
  \includegraphics[width=.9\linewidth]{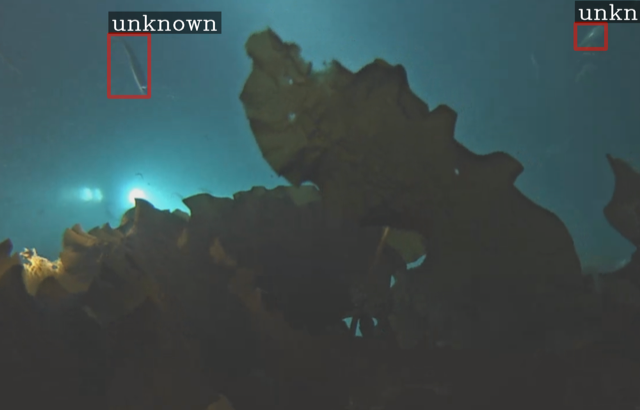}
\end{subfigure}%
\caption{\label{TemperateDatasetFigureObjectDetection} Examples from the temperate species dataset used for object detection.}
\label{fig:fig}
\end{figure*}

\subsection{Object Detection}
The object detection component is responsible for detecting the presence of fish in a video stream. The video stream can also be a live, something that limits the applicability of top level accuracy segmentation algorithms. Consequently, YOLOv3 \cite{yolov3techreport} was selected as detection algorithm. This CNN architecture provides a reasonable speed/accuracy tradeoff, and is suitable for real time implementation. The object detection takes the (live) video stream as input and outputs objects of fish without any categorization.

YOLOv3 was initialized with weights trained on ImageNet, and then further specialized by training on a new dataset. Fig.~\ref{TemperateDatasetFigureObjectDetection} shows examples from this temperate fish species detection training dataset with $619$ images containing a total of $1943$ carefully annotated fish. We deliberately designed the set up realistically for the shallow-water fish assemblage found on along the coast in Southern Norway, including the fish species most frequently observed in this ecosystem. We collected video data at several different locations, spanning depths from 1-40 meters.  We used images captured at different seasons, time-at-day (including some images captured at night) and during various weather conditions. This ensured that the dataset reflects the natural variability in visibility and light conditions. The variability is to ensure a realistic dataset as possible to ensure high precision when applied in real-life settings.

Further, note that although the detection training dataset is annotated with species, this information is not used in this stage. The object detection solely detects the presence of fish, and the categorization happens in the independent next step. The species information is used as additional data in the subsequent step. Only a fraction of Cod images are used for both detection (YOLO) and classification (CNN-SENet) training, so the datasets could be considered to be nearly non-overlapping. However, including all the temperate species classification training data in annotated form for detection should not be considered difficult, only laborious.

\subsection{Classification}
In the classification-part, two datasets were used in the test. The Fish4Knowledge dataset \cite{li2015fast} and a novel dataset with temperate species from Southern Norway, combining images from multiple surveys and field studies. Fish4Knowledge is used in pre-training of the neural network, while the temperate dataset is used in the post-training. Some differences between the datasets are: (1) The Fish4Knowledge has in addition to the fish images categorized images in trajectories, e.g. a sequence of images taken from the same video sequence or stream. (2) The temperate dataset has in addition to the other species a separate folder for male and female \textit{Symphodus melops}. Some individuals of male \textit{S. melops }have also been tracked and captured by camera multiple times. %These images have been placed in individual folders in concordance with the ID of the fish.
\subsubsection{Fish4Knowledge}
The Fish4Knowledge dataset is a collection of images, extracted from underwater videos of fish, off the coast of Taiwan. There is a total of 27230 images cataloged into 23 different species. The top 15 species accounts for 97\% of the images, and the single top species accounts for around 44\% of the images. The number of images for each species range from 25 to 12112 between the species. This creates a very imbalanced dataset. Further, the images size ranges from approximately $30\times30$ pixels to approximately $250\times250$ pixels. Another observation in the dataset, is that most of the images are taken from a viewpoint along the anteroposterior axis, or slightly tilted from that axis. In that subset of images, most of these images are from the left or right lateral side, exposing the whole dorsoventral body plan in the image. There are some images from the anterior view, but few from the posterior end. Among all the images there were not many images from the true dorsal viewpoint. Most of the selected species have a compressed body plan, e.g. dorsoventral elongate. This creates a very distinct shape when the images are taken from a lateral viewpoint. Hence, images taken from the dorsal view creates a thin, short shape. The images also have a background that is relatively light, enhancing the silhouette of the fish.
\begin{figure}[ht]
    \centering
    %\scalebox{0.75}{\includegraphics[trim=1.0cm 0.0cm 0.0cm 0.0cm]{YOLOv3_CNN_Architecture_from_a_functional_viewpoint.eps}}
    \includegraphics[width=0.5\textwidth]{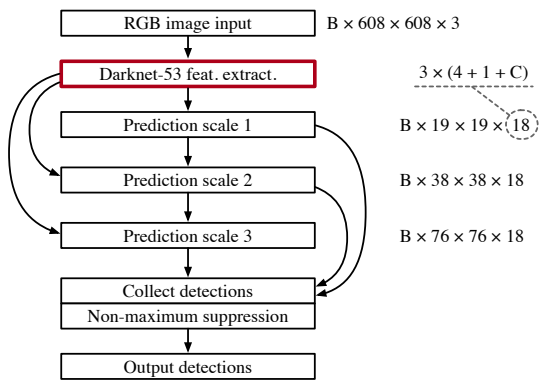}
    \caption{\label{CNN-YOLOv3Figure}A functional view of the YOLOv3 architecture.}
\end{figure}%
\begin{figure}[ht]
    \centering
    \scalebox{0.75}{\includegraphics{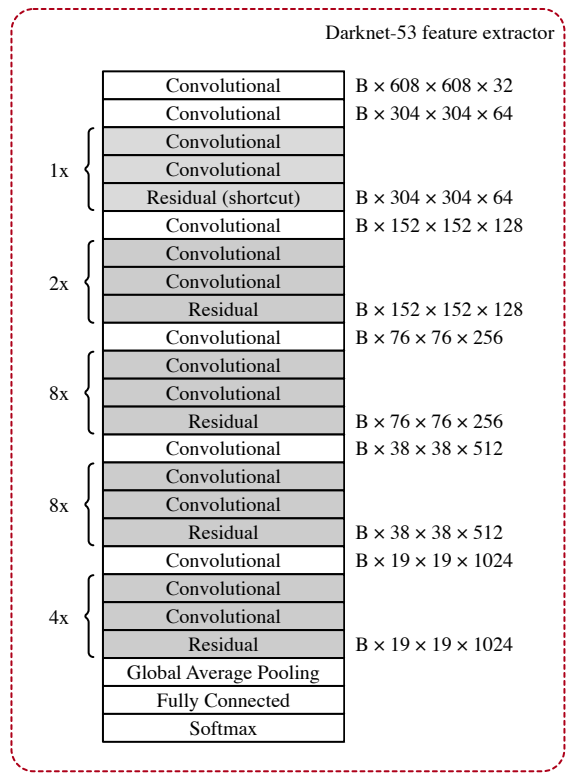}}
    \caption{\label{CNN-YOLOv3Darknet}Darknet-53 architecture with input size $608\times608\times3$ (based on \cite{yolov3techreport}).}
\end{figure}%

%\todo{add image examples from the different viewpoints}
%
\subsubsection{Temperate Fish Species}\label{TemperateFishSpecies}
\begin{figure*}[ht]
\begin{center}
\includegraphics[width=\linewidth,keepaspectratio]{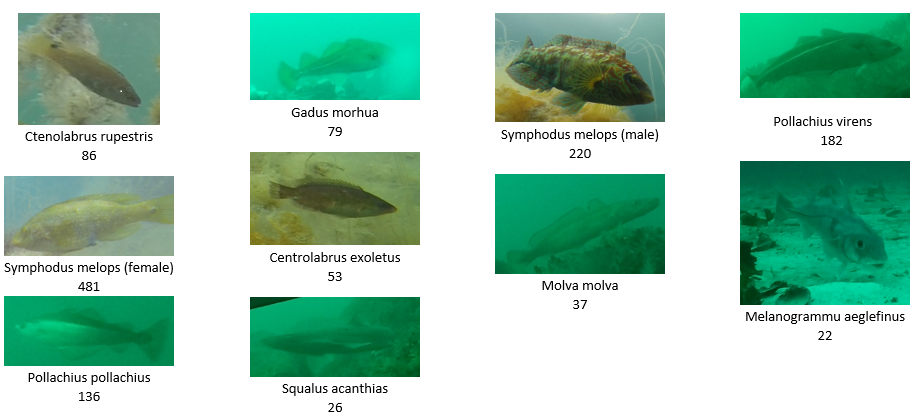}
\end{center}
\caption{\label{TemperateDatasetFigure}Example images and distribution of the temperate species dataset used for classification.}
\end{figure*}%
The temperate dataset is a collection of images from some of the most abundant fish species in coastal areas of Northern Europe. Video recordings from GoPro cameras (HERO4-7+Black) were obtained at three different locations from south to western Norway between 2014 and 2019. In western Norway, Austevoll, the cameras were deployed at 2-5 meters of depth around small reef sites used as breeding sites for many wrasse fishes. The species identified from these videos were \textit{Ctenolabrus rupestris}, \textit{Centrolabrus exoletus} and  \textit{S. melops}. In \textit{S. melops}, most males build nests to care for eggs and are colourful and easily distinguished from the brown coloured females \cite{HalvorsenEtAl2016MaleBiased}. However, a minority of the males are visually indistinguishable from females and use this camouflage to sneak on other males’ nest to steal fertilization \cite{HalvorsenEtAl2017Sex-And-Size-Selective}. Because of the morphological appearances of the different sexes, nest-building males are labelled as ``males" in the dataset, whereas females and sneaker males are labelled as ``females". Two of the wrasse species (\textit{Ctenolabrus rupestris} and \textit{S. melops}) have high commercial importance as they are used as cleaner fish in the aquaculture industry. In the south-eastern Norway, county of Agder, and mid-western Norway, county of Trøndelag, stereo baited remote underwater video (stereo-BRUV) rigs were deployed at 8-35 meters of depth at various shallow coastal habitats. From these videos, we extracted frames showing species from the family Gadidae : \textit{Gadus morhua, Pollachius virens, Pollachius pollachius, Molva molva, and Melanogrammus aeglefinus}, all with commercial importance. Additionally, some images shows \textit{Squalus acanthias}, a shark classified as vulnerable globally and critically endangered in the Northeast Atlantic by the IUCN red list of threatened species~\cite{Fordham2016}.

The temperate dataset has a higher image noise level and more variability compared with the Fish4Knowledge dataset, such as differences in depth, visibility and habitat, and orientation of the fish and distance between camera and fish. This secured a high variability in pictures of each species and a natural representative for observations in wild, but it is also expected to reduce the classification accuracy. Furthermore, a single video frame usually contained more than one fish (e.g., the same species, different species). All videos were recorded in full HD resolution of $1920\times1080$ pixels with default settings. Fig.~\ref{TemperateDatasetFigure} illustrates samples of the dataset.

%
% Dataset size: 86+79+220+182+481+53+37+22+136+26 = 1322; 
% males: 220/1322 ~= 17%
% females and sneakers: 481/1322 ~= 36%
\section{Object Detection and Classification}\label{structure}
In contrast to the available literature, we have separated object detection from classification. This separation allows for both separate training data for fish detection and species classification, and different level of validity in the training data. It also allows for a much more fine-grained classification of species independent from detecting the fish.

\subsection{Fish Detection}\label{SectionFishDetection}%
Fish are detected independent from species recognition through object detection using YOLOv3. YOLO is a state-of-the-art object detector, originally designed for combined detection and classification. Only the detection part is used in this work. YOLO is efficient, and provides relatively high accuracy at the same time as being moderately computationally expensive~\cite{yolov1techreport,yolov3techreport}. Combined with the speed and accuracy of CNN-SENet for species classification, this should enable real time applications even on embedded devices such as NVIDIA Jetson AGX Xavier and Intel Movidius Myriad variants.

YOLOv3 is configured to detect and classify only one class ($C=1$), namely ``fish", and use an input image of dimension $608\times608$ with three color channels in RGB order. Default initial values for the nine object detecion bounding box priors were used (width$\times$height): $10\times13$,  $16\times30$,  $33\times23$,  $30\times61$,  $62\times45$,  $59\times119$, $116\times90$, $156\times198$ and $373\times326$. These values are recommended for the COCO dataset. By inspection, the fish dataset will contain approximately the same kind of variations in object sizes and orientations, with both horizontally and vertically oriented objects. If we intended to use this algorithm in a \emph{structured} environment, where for example, all the fish were expected to swim through an apparatus, it would have been interesting to explore a prior distribution favoring slender horizontally oriented rectangular boxes. Note that sizes are given in pixels, relative to the scaled version of any given image.

When training the network, a batch size configuration $B$ of $64$ and $8$ subdivisions was configured. The number of subdivisions required was found experimentally and is dependent on the available training hardware (GPU RAM). Four NVIDIA V100 GPUs in a DGX-2 computer were used. Convolutional weights were initialized with weights pre-trained on ImageNet~\cite{2014OlgaImagenet} data. Next, the training process was started using a single GPU for $4000$ iterations as ``burn-in". As a consequence of the number of GPUs available, and the relatively small dataset, the default Darknet YOLOv3 learning rate was reduced by a factor of $0.25$ to $0.00025$ during this training phase. The effect of different learning rate is visible in Fig.~\ref{iou} as increased variability from batch 4000. After ``burn-in" the training was stopped and then restarted from saved weights using four GPUs. Training was configured to run $50000$ iterations in total. This is equivalent to approximately $7000$ epochs given a batch size of 64 and 434 training images. The step yielding the best mean average precision (mAP@50) is selected for detection use. Both the original ``Darknet" framework from the YOLOv3 authors and an extended, forked, version was used for running the experiments\footnote{\url{https://github.com/AlexeyAB/darknet}}.

\subsection{Species Classification}
\begin{figure}[ht]
    \centering
    \scalebox{0.75}{\includegraphics{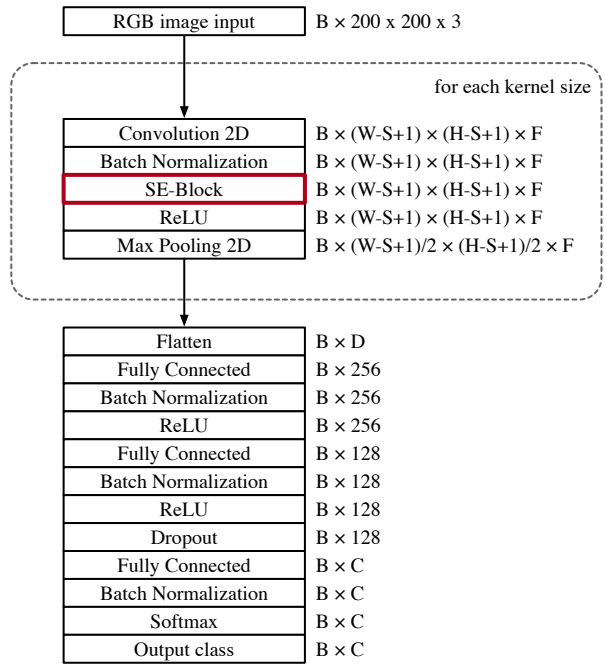}}
    \caption{\label{CNN-SENetFigure}CNN-SENet architecture.}
\end{figure}

\begin{figure}[ht]
    \centering
    \scalebox{0.75}{\includegraphics{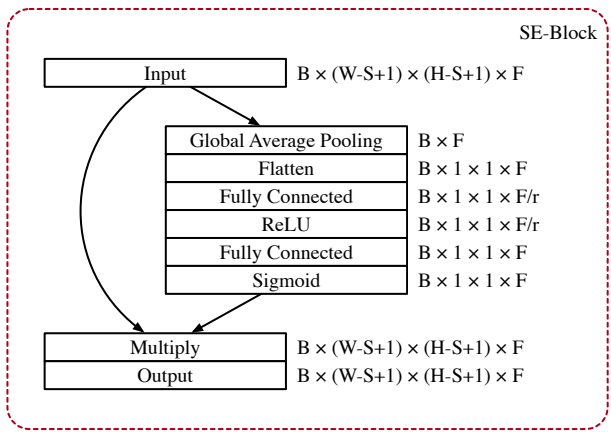}}
    \caption{\label{SE-Block}Squeeze-and-Excitation block.}
\end{figure}

The species of the fish is identified by classification using a Convolutional Neural Network with an added squeeze and excitation (SE) -- using the CNN-SENet structure.
A CNN-SENet is an architectural element that re-calibrates channel wise-feature responses adaptively~\cite{HuLiGang2017}.
The architecture of the CNN-SENet, depicted in Fig.~\ref{CNN-SENetFigure}, is configured with the following parameters. Image size in height ($H$), width ($W$) and depth channels; the number of learnable filters ($F$); the batch size ($B$) (default 16), the filter size ($S$), and reduction ratio ($r$) as described in~\cite{HuLiGang2017}. Lastly, the number of fish species classifications needs to be added, as parameter $C$.
The input layer takes an image of size $200\times200$ with a depth of 3 color channels, R, G, and B. The output is batch normalized before entering the Squeeze-and-Excitation function, called SE block, depicted in Fig.~\ref{SE-Block}.
The SE block performs a feature re-calibration through the (1) squeeze operation preventing the network from becoming channel-dependent. This exploits contextual information outside the receptive field and is achieved by doing global average pooling on each input channel before reshaping, and (2) the excitation operation that utilizes the output from the squeeze function by fully capture channel-wise dependencies. This is achieved by the two fully-connected (FC) layers sandwiching the reduction layer, and finally, a sigmoid activation layer. Before exiting the SE block, the output from the excitation function is multiplied with the original batch normalized output. This multiplied output is then added to a ReLU layer performing an element-wise activation function, rendering the dimension size unchanged. The output is then sent to a Max Pooling layer, which uses a $2\times2$ filter to reduce and re-size the height and width spatially, rendering output of $98\times98\times32$. This core portion of the network is stacked to the size of the kernel size, in this case, the size of five. The first iteration has a convolutional layer of 32 filters in $5\times5$. The second and third have 64 filters in $3\times3$, the forth 128 filter in $2\times2$,  and the fifth 256 filters in $2\times2$, with all layers applying a horizontal and vertical stride of 1.

Furthermore, the network has 3 FC layers. The first, with 256 neurons, takes the output from the last convolutional layer that is first flattened. The output is then batch normalized before sent to the second FC layer, with 256 neurons. A reduction function is applied after the output from the FC layer is batch normalized. Before entering the last FC layer, with $C$ neurons, a dropout layer of 50\% is applied. The final layer, softmax, applies a classifier function to obtain the probability distribution for each class per input image, using a categorical cross-entropy with the Adam optimizer \cite{kingma2014adam}.

In CNN-SENet, there are specific parameters that need to be configured, including dropout percentage, learning rate, and batch normalization, that are discussed presently. The parameters are configured based on the trial-and-error method. For the dropout percentage, clearly, the higher the dropout, the more the information is lost during training because forward- and back-propagation are carried out only on the remaining neurons after dropout is applied. Different percentages of the dropout are tested, and 50\% is configured in this study due to the better overall performance achieved. The learning rates when using the Adam optimizer should be tuned to further optimize the network. After numerous trials, the learning rate is configured as 0.001 without decay. For batch normalization, it has been tested, and the results with batch normalization are slightly better than without it. In more detail the accuracy of the testing set without batch normalization is 98.35\%, while the accuracy with batch normalization is 99.27\%. With the above parameters, the model trains faster and has a higher validation accuracy, which concludes the architecture of CNN-SENet. 

%The pre-training process includes only images from the Fish4Knowledge dataset. No filters have been applied in this process. Since all images need to have the same dimension for CNN to work, the images are re-sized to 200x200 pixels. Further,  
To compare CNN-SENet with DeepFish, Table~\ref{SENetVsDeepFishTable} illustrates the main differences between the two. Clearly, CNN-SENet has a more sophisticated structure than DeepFish.
\begin{table*}[ht!]
\centering
\caption{Differences between CNN-SENet and DeepFish.}
\begin{tabular}{|l|l|l|}
\hline
 & \textbf{CNN-SENet} & \textbf{DeepFish} \\ \hline
\textbf{Image Size} & $200\times200$ & $47\times47$ \\ \hline
\textbf{Testing Samples} & 4126 & 3098 \\ \hline
\textbf{\begin{tabular}[c]{@{}l@{}}Network\\Architecture\end{tabular}} & Basic with SE blocks & Basic \\ \hline
\textbf{Classifier} & Softmax & SVM \\ \hline
\textbf{\begin{tabular}[c]{@{}l@{}}Convolutional\\Layers\end{tabular}} & 5 & 3 \\ \hline
\end{tabular}\label{SENetVsDeepFishTable}
\end{table*}% \pbox{25cm}{ \hline
\section{Experiments, Results and Discussion}\label{experimentalResults}
The proposed approach was verified in a two-step approach using separate experiments for fish detection and classification. First performance of fish detection was assessed, then the performance of fish classification.
\subsection{Fish Detection}
Localization of individual fish in each video stream image occurs with the YOLOv3 based object detector described in Section~\ref{SectionFishDetection}. Detection accuracy is measured using Intersection over Union (IoU) -- Jaccard index. This is a measure of overlap between two sets, and a widely used measure for verification of object detection and segmentation algorithms. The approach reaches an average IoU of 0.6802, and an IoU per class 0.9934. The latter number means that a tiny percentage of false objects consisting of mere background was erroneously detected as fish.

\begin{figure}[ht]
    \centering
    \includegraphics[width=1.0\linewidth]{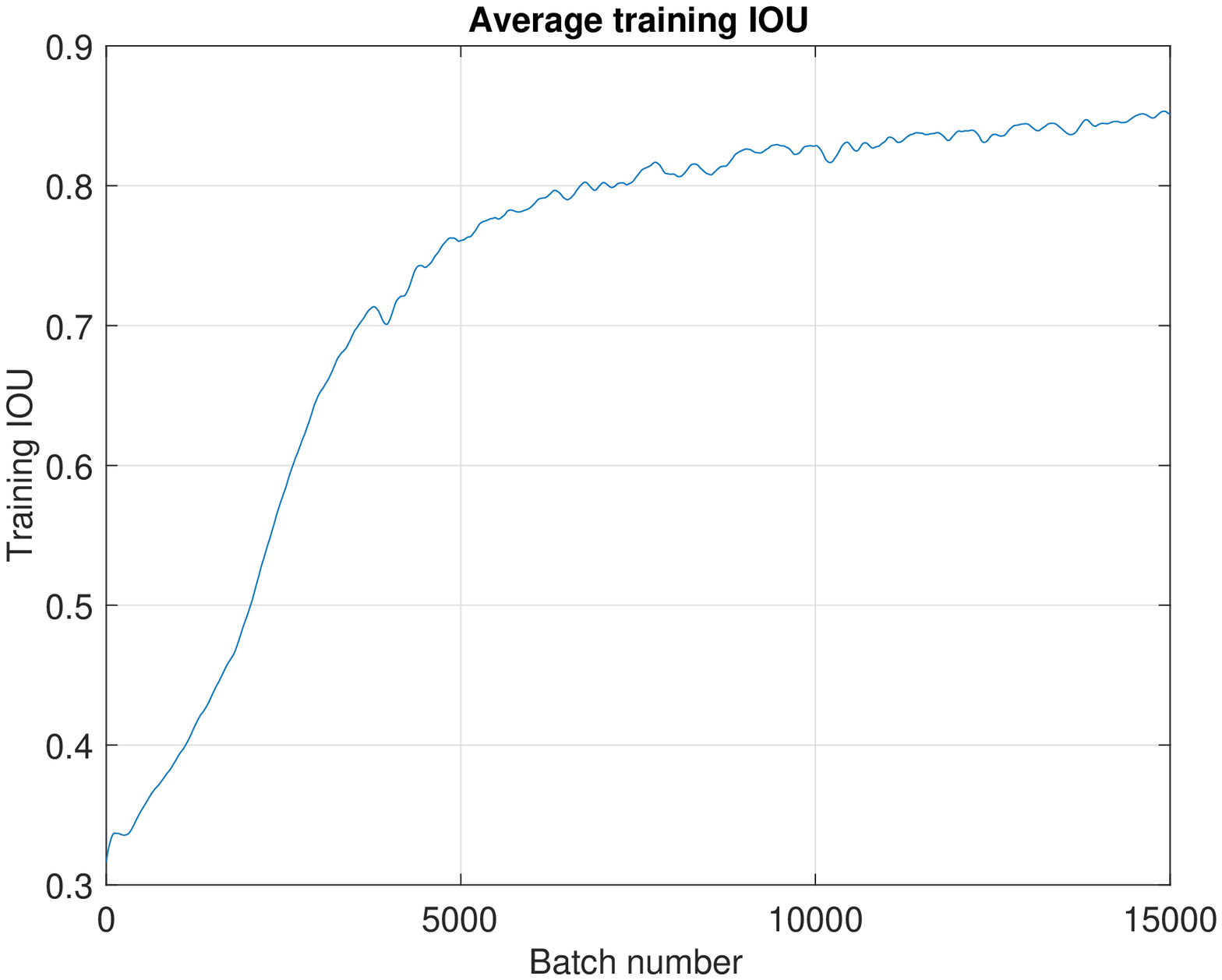}
    \caption{\label{iou}Training Intersection over Union (IoU) with moving average.}
\end{figure}

\begin{figure}[ht]
    \centering
    \includegraphics[width=1.0\linewidth]{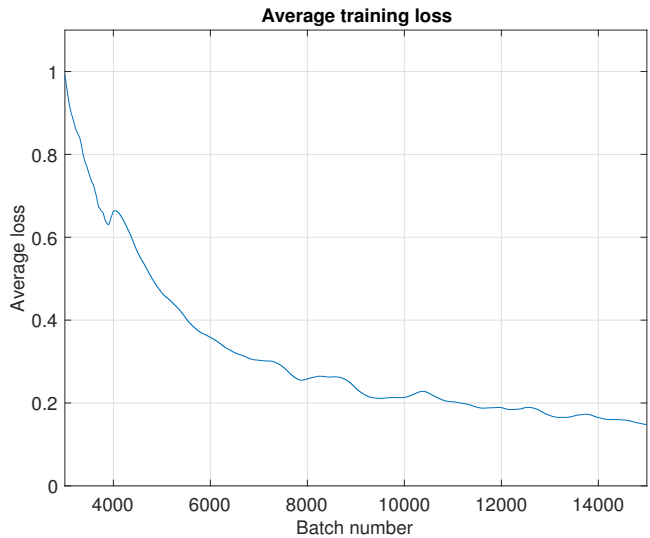}
    \caption{\label{yolov3_training_total_loss}Total training loss after moving average filter.}
\end{figure}

\begin{figure}[ht]
    \centering
    \includegraphics[width=1.0\linewidth]{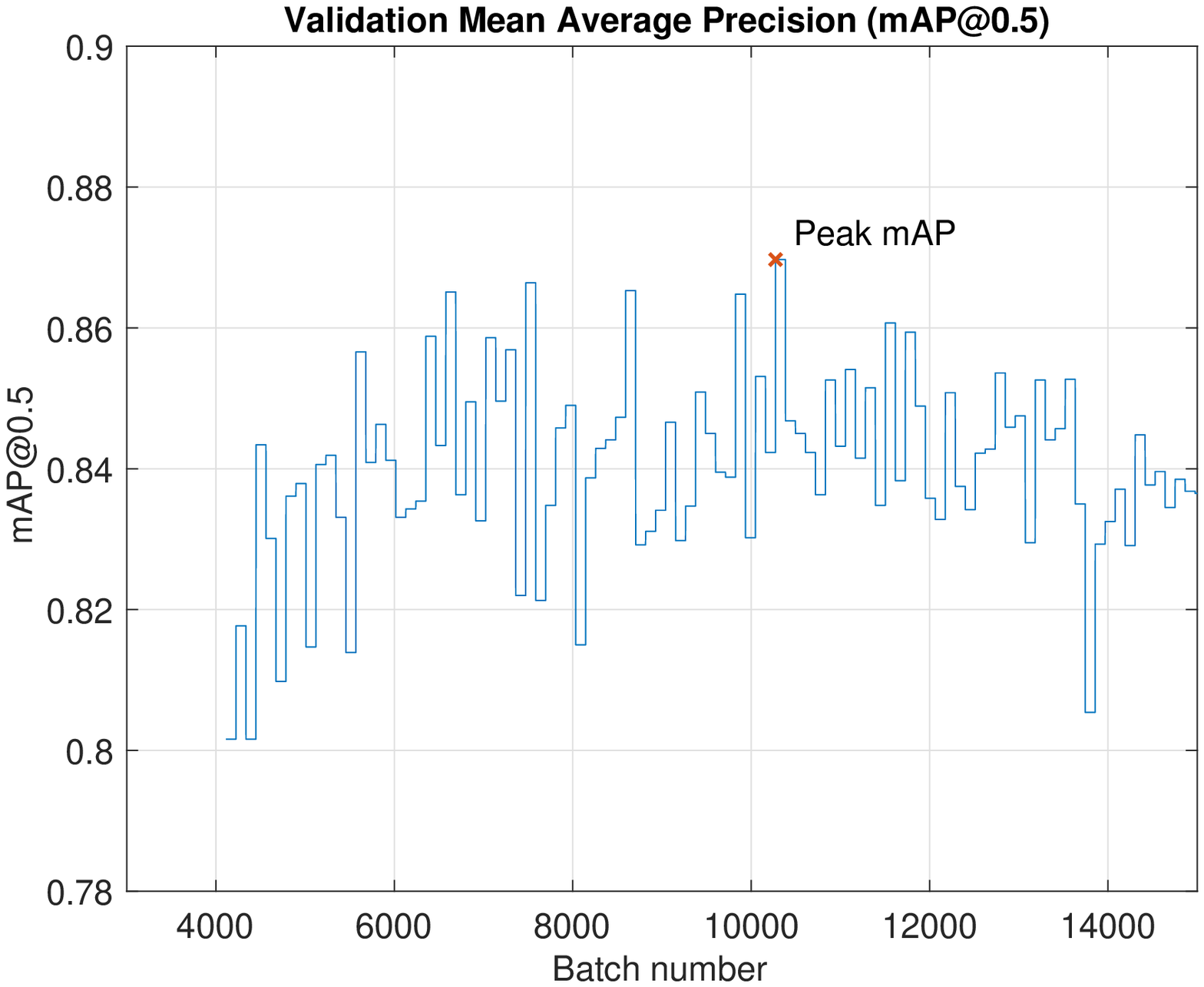}
    \caption{\label{yolov3_validation_mAP_peak}Mean Average Precision (mAP) with peak value 86.97 \% at batch 10273.}
\end{figure}

The dataset for this experiment was randomly split in a 70\% for training and 30\% for verification. Fig. \ref{iou} shows IoU per epoch for the latter. Fig. \ref{yolov3_training_total_loss} and \ref{yolov3_validation_mAP_peak} show the training loss and mean average precision, respectively. The precision peaks at 86.96\%. 

\begin{figure*}[t!]
\begin{subfigure}{.5\textwidth}
  \centering
  \includegraphics[width=.9\linewidth]{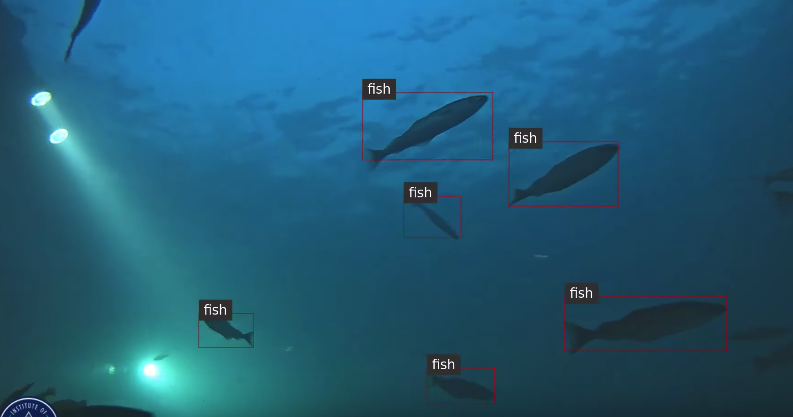}
  \caption{Day time correct detection}
\end{subfigure}%
\begin{subfigure}{.5\textwidth}
  \centering
  \includegraphics[width=.9\linewidth]{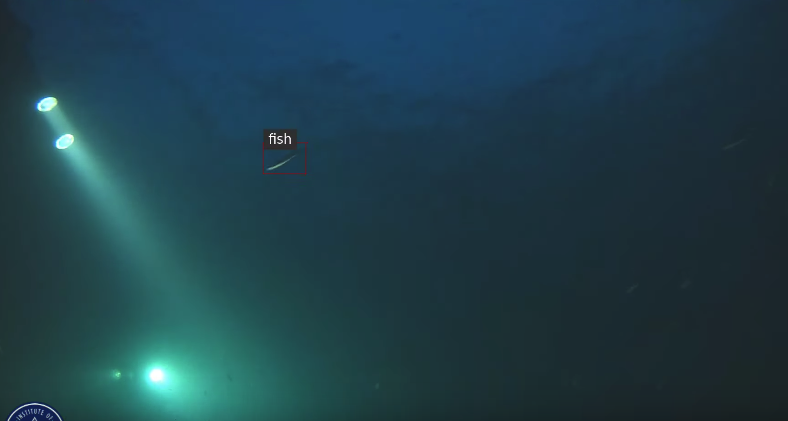}
  \caption{Dark late evening correct detection}
\end{subfigure}%
\\
%\end{figure*}
%\begin{figure*}
\begin{subfigure}{.5\textwidth}
  \centering
  \includegraphics[width=.9\linewidth]{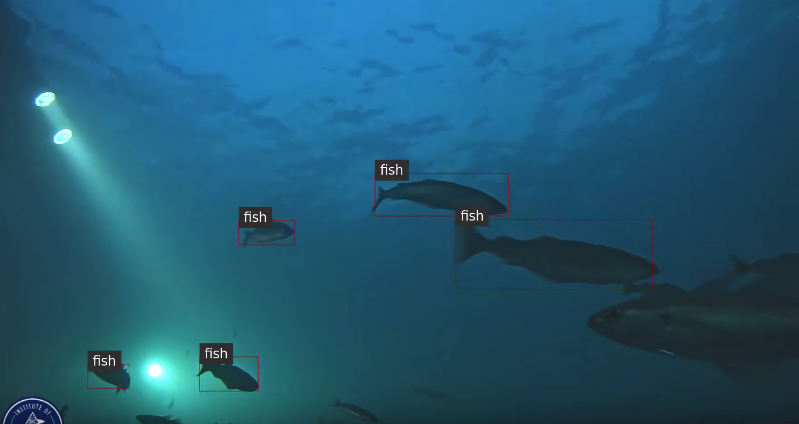}
  \caption{Day time partially correct detection}
\end{subfigure}%
\begin{subfigure}{.5\textwidth}
  \centering
  \includegraphics[width=.9\linewidth]{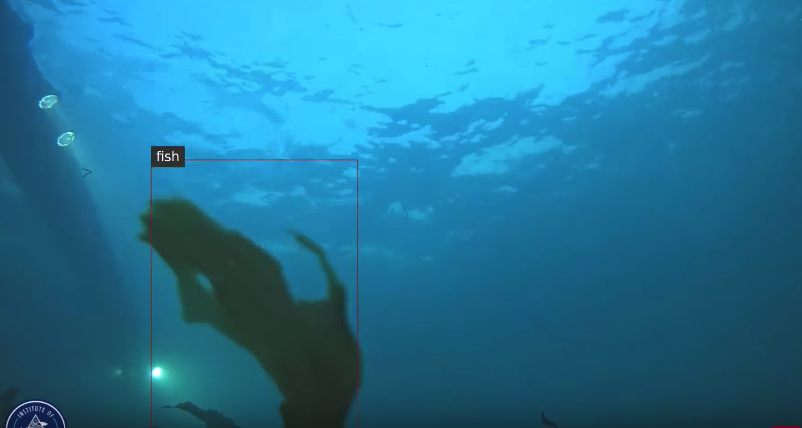}
  \caption{Fail case}
\end{subfigure}%
\caption{\label{Exampleresults}Three Sample frames of correct fish detection, and one erroneous case, extracted from underwater video-stream.}
\end{figure*}

The validity of our approach is further confirmed in a different setting than the training data. This verification is part of a live stream from an underwater camera located near a semi-submerged restaurant in southern Norway, and which provide highly variable lighting conditions, and different camera angles not part of our training data\footnote{A recording of real-time detection is available at \url{https://www.youtube.com/watch?v=bZMJEIWo-rQ&t=4298s}}. Despite the radically different scenarios, the proposed method is still able to detect fish correctly with very high accuracy.
Fig.~\ref{Exampleresults} shows samples from the live stream recording. Three of the examples show fish which are correctly detected, and one failed case. The first case in Fig.~\ref{Exampleresults} shows the standard case during day time, the second shows fish detected during dark evenings with artificial light, and the third case shows most of the fish detected while the fish in the corner are wrongly ignored. In the last occurrence, seaweed is detected as fish. 


\subsection{Species Classification}
Classification of species is done by categorizing fish identified in the object detection. Accuracy and performance of the new fish classification CNN-SENet are quantified and compared with the state-of-the-art networks represented by Inception-V3, ResNet-50, and Inception-ResNet-V2. Additionally, a simplified version of the CNN-SENet, without the Squeeze-and-Excitation blocks, is included to explore how the spatial relationship between fish image colors and other feature layers affect results \cite{HuLiGang2017}.

Three different experiments were performed. Pre-training with Fish4Knowledge, post-training with the new temperate Fish Species dataset described in Subsection~\ref{TemperateFishSpecies} and post-training with an extended version of the new dataset using image augmentation techniques. For all three experiments, the relevant dataset was divided into 70\% training images, 15\% validation images, and 15\% testing images. Both training and validation images are integral parts of the training process, while the testing images were kept out-of-the-loop for independent verification of the ``end product".

All benchmarked networks are trained for 50 epochs with images adapted to their input image size of $200\times200$ RGB pixels, with the notable exception of the $299\times299$ RGB pixels required by Inception-ResNet-V2.
\subsubsection{Pre-training}
\begin{figure*}[ht!]
\begin{center}
\includegraphics[width=.8\linewidth]{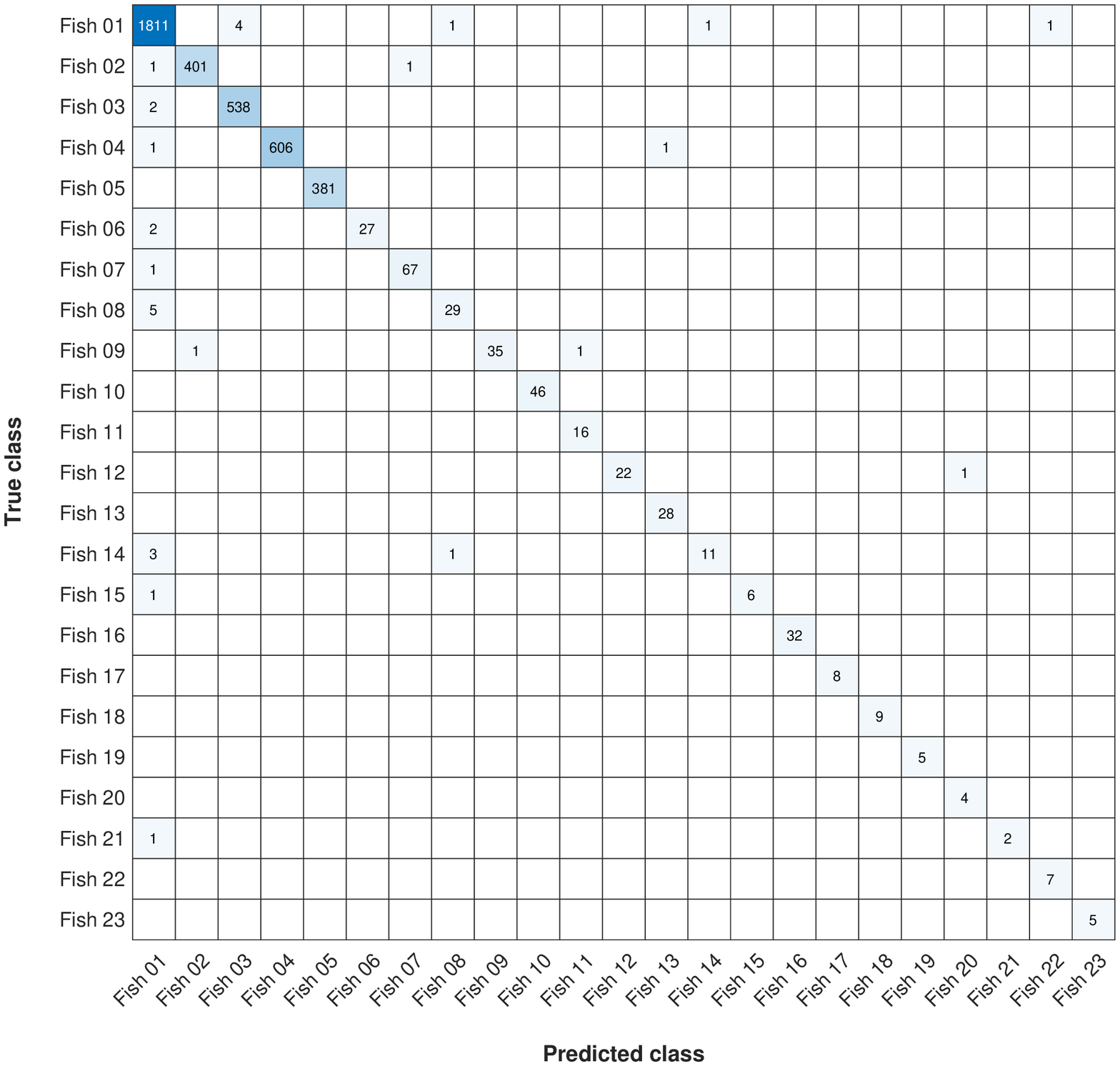} % trim=0cm 0.8cm -2.2cm 1.0cm
\end{center}
\caption{\label{ConfusionPreFigure} Confusion matrix for Fish4Knowledge dataset pre-training with CNN-SENet.}
\end{figure*}
Pre-training was performed using a dataset consisting of 19149 Fish4Knowledge images, with an additional 4126 images for verification and 4126 images reserved for testing. The selected training configuration consists of a single run with 50 training epochs and a batch size of 16. Results from pre-training are evaluated using weights from the epoch with the highest validation accuracy, and not necessarily the final epoch. 
\begin{table*}[ht!]
\centering
\caption{Testing accuracy and time per epoch on pre-training.}
\begin{tabular}{|c|c|c|}
\hline
\textbf{Network}                     & \textbf{Testing Accuracy}		& \textbf{Time One Epoch} \\ \hline
\textbf{Inception-V3}   &    99.18\%    &923 s                \\ \hline
\textbf{ResNet-50}      &   98.86\%     &646 s                 \\ \hline
\textbf{Inception-ResNet-V2}    &    98.59\% & 2221 s                  \\ \hline
\textbf{CNN-SENet}      &    99.27\%  & 197 s                 \\ \hline
\textbf{\begin{tabular}[c]{@{}l@{}}CNN-SENet without\\ Squeeze-and-Excitation\end{tabular}}      &   99.15\%    & 159 s                    \\ \hline
\end{tabular}\label{PreTrainingAccuracyTable}
\end{table*}
\subsubsection{Post-training}
\begin{figure*}[ht]
\begin{center}
\includegraphics[width=.55\textwidth]{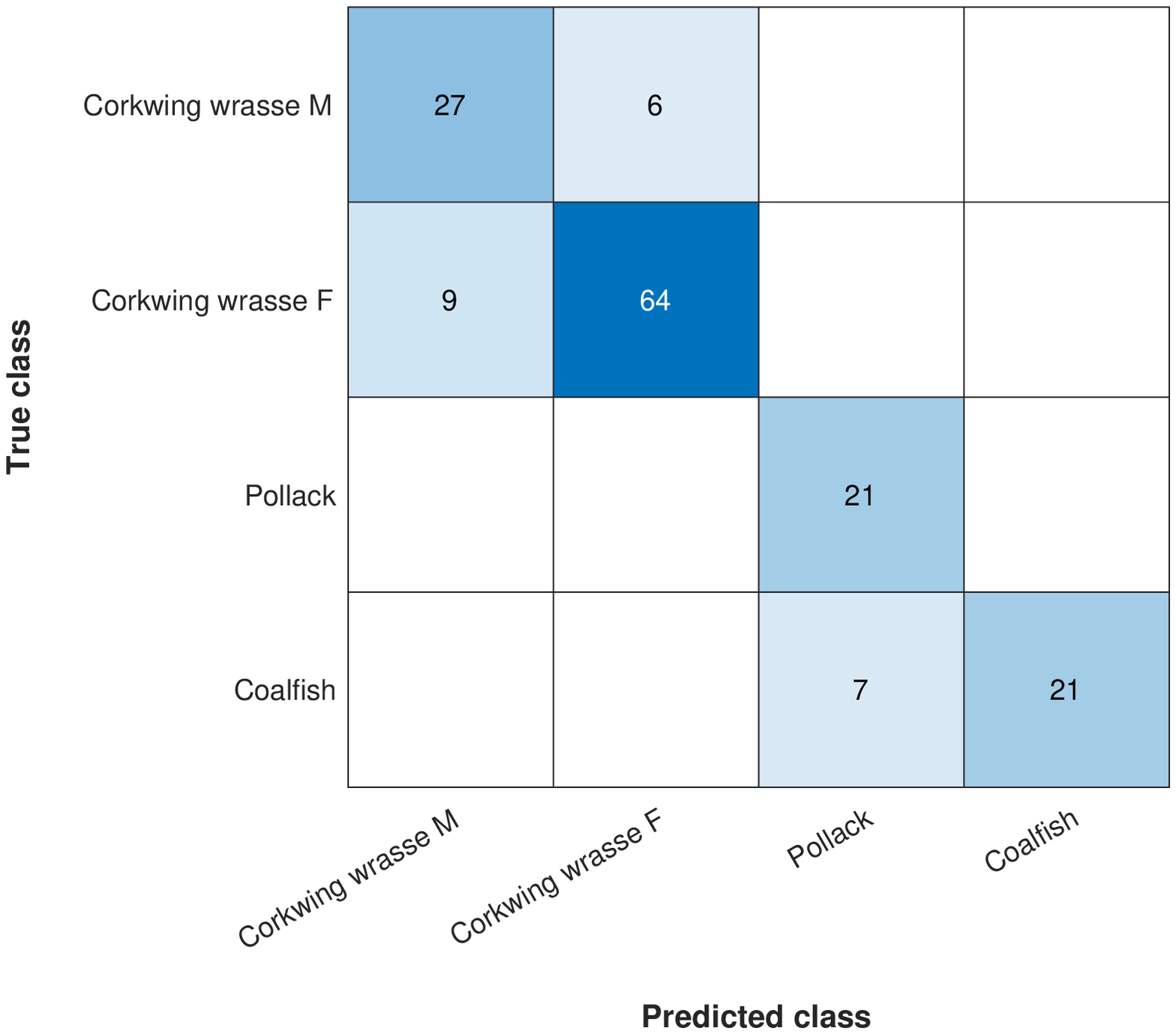} % , trim=0cm 0.8cm -2.1cm 0cm
\end{center}
\caption{\label{ConfusionPostFigure} Confusion matrix for temperate dataset post-training with CNN-SENet.}
\end{figure*}
Post-training was performed using 712 images of four fish classes from the temperate fish species dataset described in Section~\ref{TemperateFishSpecies}. An additional 155 images were used for verification during training, and a subset of 155 images of the same classes were reserved for testing. Corkwing wrasse (male), Corkwing wrasse (female), Pollach, and Coalfish were selected for the experiment as a reasonable number of images of different individuals under varying conditions were available for these species.

The post-training process consists of 50 epochs and a batch size of 8. The batch size was reduced, compared to pre-training, to compensate for the relatively small number of available temperate fish images. Weights from the pre-training step are loaded before initiating post-training, and post-training accuracy is evaluated using the weights from the final epoch.

The rationale for this post-training method is to make use of the more or less generic fish identification features learned from the large Fish4Knowledge dataset. Post-training will then start with the network in a ``fish-class-sensitive" state and proceed by learning specific features of the temperate species on top of this.

Fish4Knowledge consists of images of 23 different classes. The selected subset of the temperate dataset consists of 4 classes. To prepare the loaded pre-trained model for post-training, the last fully connected (FC) layer with 23 output neurons, suitable for 23 fish classes, is replaced with a similar layer with four output neurons.
\begin{table*}[ht]
\centering
\caption{Average testing accuracy over 10 runs and time per epoch on post-training.}
\begin{tabular}{|c|c|c|}
\hline
\textbf{Network}                     & \textbf{Testing Accuracy}		& \textbf{Time One Epoch} \\ \hline
\textbf{Inception-V3} &    85.42\%         & 33 s              \\ \hline
\textbf{ResNet-50}    &    82.39\%        & 47 s               \\ \hline
\textbf{Inception-ResNet-V2} & 78.84\%  	& 91 s                \\ \hline
\textbf{CNN-SENet}       &    83.68\%       & 9 s           \\ \hline
\textbf{\begin{tabular}[c]{@{}l@{}}CNN-SENet without\\ Squeeze-and-Excitation\end{tabular}}       &    82.32\%       & 7 s           \\ \hline
\end{tabular}\label{PostTrainingAccuracyTable}%
\end{table*}
\subsubsection{Post-training with Image Augmentation}\label{ImageAugmentationExperiment}
Data augmentation techniques in machine learning aims at reducing overfitting problems by expanding a dataset (base set) by introducing label-preserving transformations. For an image dataset, this means that transformed copies of the original images in the base set are produced. These additional training data enable a network under training to learn more generic features by reducing sensitivity to augmentation operations that transform the image but not severely the characterizing visual features of, for example a fish \cite{KrizhevskyImagenet2012ImageAugmentation}. 

The main algorithm flow is the same as for the post-training version, but the dataset was expanded by using the following transformation operations. Images are rotated randomly within a specific range, according to a uniform distribution. Images are vertically and horizontally shifted a random fraction of the image size. Scaling and shearing transformations are applied randomly, and lastly, half of the images are flipped horizontally.

\subsection{Results}

\subsubsection{Pre-training}
Results from pre-training on Fish4Knowledge are presented in Table~\ref{PreTrainingAccuracyTable}. The testing accuracy is on par with or exceeds the level of accuracy achieved with previous state-of-art solutions described in Section~\ref{Introduction}.

CNN-SENet with Squeeze-and-Excitation achieves 99.15\% test accuracy, almost identical results as the Inception-V3 algorithm when it comes to accuracy. However, the run time for each epoch is roughly three times larger for Inception-V3. The training-runtime is expected to be reflected in prediction. CNN-SENet without Squeeze-and-Excitation is faster than the SE-version, but also slightly less accurate during these tests.

Inception-ResNet-V2 achieves the lowest test accuracy and also the highest time consumed for each epoch during training. The required input image size is $299\times299$, compared to $200\times200$ for the other networks under test. As the required resolution is higher than the resolution of most Fish4Knowledge images, the necessary upscaling process may negatively affect accuracy. Additionally, the larger input size also dramatically increases the computational complexity and leads to a longer time on each epoch.

A confusion matrix for the CNN-SENet pre-training run is included, as shown in Fig.~\ref{ConfusionPreFigure}. Fish 01 seems to attract more wrong predictions than the other species. The reason for this is unknown, but the imbalance in the dataset could explain some of the behavior, as the ability to learn Fish 01 will be more rewarding during training as it occurs more frequently.

\subsubsection{Post-training with and without image augmentation}
Results from the post-training experiment indicates that this is a more challenging image recognition task. Without image augmentation, the highest average testing accuracy achieved was 85.42\% using the Inception-V3 CNN algorithm as, listed in Table~\ref{PostTrainingAccuracyTable}. CNN-SENet performance is a few percent below, but with a significantly better training time for each epoch. All bench-marked algorithms show significantly reduced accuracy compared to the results from pre-training. The temperate species dataset used for post-training is challenging, in the sense that it contains few images overall. The dataset also consists of pictures of fish under low visibility conditions and situations where the fish silhouette is not always prominent.

Image augmentation, as described in Section~\ref{ImageAugmentationExperiment}, improves the results for post-training for all benchmarked algorithms, as shown in Table~\ref{PostTrainingWithImageAugmentationAccuracyTable}. The ResNet-50 network reaches just above 90\% testing accuracy. CNN-SENet accuracy increases approximately four percentage points compared to post-training without image augmentation. The training time for each epoch does not change notably using image augmentation, so the metric was omitted from Table~\ref{PostTrainingWithImageAugmentationAccuracyTable}.
\begin{table}[ht]
\centering
\caption{Average testing accuracy over 10 runs on post-training with image augmentation.}
\begin{tabular}{|c|c|}
\hline
\textbf{Network}                     & \textbf{Testing Accuracy}	\\ \hline
\textbf{Inception-V3} &    88.45\%      \\ \hline
\textbf{ResNet-50}    &    90.20\%         \\ \hline
\textbf{Inception-ResNet-V2} & 82.39\%       \\ \hline
\textbf{CNN-SENet}       &    87.74\%   \\ \hline
\textbf{\begin{tabular}[c]{@{}l@{}}CNN-SENet without\\ Squeeze-and-Excitation\end{tabular}}       &    83.55\%       \\ \hline
\end{tabular}\label{PostTrainingWithImageAugmentationAccuracyTable}
\end{table}
\section{Conclusions}\label{conclusion}
In this study, we implemented an in-depth deep learning-based approach for temperate fish detection and classification. YOLOv3 has been used for detection purposes, and CNN-SENet has been adopted for classification. The experimental results show that the YOLOv3 technique can successfully detect an individual fish in different complex environmental conditions. The object detection approaches a mean average precision of 86.96\%, and the CNN-SENet architecture achieves the state-of-the-art accuracy of 99.27\% on the Fish4Knowledge dataset without any data augmentation or image pre-processing. For temperate fish, the obtained average accuracy is 83.68\%. The lower accuracy can be explained by the comparatively smaller temperate species dataset combined with high variation in image data. The detection algorithm was also tested successfully in real-time on a live 25 FPS Full HD underwater video stream. In short, we show that our proposed deep learning approach is a powerful and useful tool for the automatic analysis of fish species. It has a high potential to release the burden on scientists working with the study of videos and pictures from underwater ecosystems.

\bibliographystyle{ieeetr}
\bibliography{_bibliography} % KMK: Now using bibtext to add new items (consider replacing with bibtex output again in final stage).

\end{document}